\documentclass{article}

\usepackage[preprint]{neurips_2025}

\usepackage[utf8]{inputenc}
\usepackage[T1]{fontenc}
\usepackage{hyperref}
\usepackage{url}
\usepackage{booktabs}
\usepackage{amsfonts}
\usepackage{amsmath}
\usepackage{nicefrac}
\usepackage{microtype}
\usepackage{xcolor}
\usepackage{graphicx}
\usepackage{subcaption}
\usepackage{multirow}
\usepackage{pgfplots}
\usepackage{listings}
\usepackage{tikz}
\usetikzlibrary{shapes,arrows,positioning,fit,backgrounds}

\pgfplotsset{compat=1.17}

\newcommand{\corecraft}{\textsc{Corecraft}}

\lstdefinestyle{jsonstyle}{
    basicstyle=\ttfamily\small,
    breaklines=true,
    frame=single,
    backgroundcolor=\color{gray!10},
    keywordstyle=\color{blue},
    stringstyle=\color{red!70!black},
    commentstyle=\color{green!50!black},
    showstringspaces=false,
    tabsize=2
}

\title{The Hierarchy of Agentic Capabilities: \\
Evaluating Frontier Models on Realistic RL Environments}

\author{
  Logan Ritchie\thanks{Correspondence to: \texttt{loganritchie@surgehq.ai}} \quad
  Sushant Mehta \quad
  Nick Heiner \quad
  Mason Yu \quad
  Edwin Chen \\[1ex]
  \textbf{Surge AI}
}

\begin{document}

\maketitle

\begin{abstract}
The advancement of large language model (LLM) based agents has shifted AI evaluation from single-turn response assessment to multi-step task completion in interactive environments. We present an empirical study evaluating frontier AI models on 150 workplace tasks within a realistic e-commerce RL environment from Surge. Our analysis reveals an empirically-derived \emph{hierarchy of agentic capabilities} that models must master for real-world deployment: (1) tool use, (2) planning and goal formation, (3) adaptability, (4) groundedness, and (5) common-sense reasoning. Even the best-performing models fail approximately 40\% of the tasks, with failures clustering predictably along this hierarchy. Weaker models struggle with fundamental tool use and planning, whereas stronger models primarily fail on tasks requiring contextual inference beyond explicit instructions. We introduce a task-centric design methodology for RL environments that emphasizes diversity and domain expert contributions, provide detailed failure analysis, and discuss implications for agent development. Our findings suggest that while current frontier models can demonstrate coherent multi-step behavior, substantial capability gaps remain before achieving human-level task completion in realistic workplace settings.
\end{abstract}

\section{Introduction}

The year 2025 has witnessed the transition of AI systems from conversational assistants to more autonomous agents capable of executing actions in real-world environments \citep{xi2023rise,wang2024survey}. This evolution raises an important question: \emph{how much (economically) useful work can AI agents actually perform?}

Addressing this question requires evaluation paradigms that go beyond traditional benchmarks. Although assessments such as MMLU \citep{hendrycks2021measuring} and HellaSwag \citep{zellers2019hellaswag} measure static knowledge and reasoning, they fail to capture the dynamic, multi-step nature of real-world tasks. Similarly, single-turn instruction-following evaluations \citep{zhou2023instruction} cannot measure an agent's ability to recover from errors, adapt to unexpected situations, or maintain coherent behavior across extended interactions.

This gap has motivated interactive evaluation environments where AI agents act through tool use and are assessed on realistic multi-step tasks \citep{zhou2024webarena,xie2024osworld,mialon2024gaia,xu2024theagentcompany}. These environments enable evaluations of agentic capabilities that static benchmarks cannot capture. Recent empirical studies of production agents reveal that despite widespread deployment in various industries, reliability remains the primary challenge, with practitioners deliberately constraining agent autonomy to maintain operational stability \citep{pan2025map}.

We present an empirical study of frontier AI models evaluated within an RL environment \corecraft{}, Inc., simulating an online retailer of high-performance PC components and custom builds. Models assume the role of customer support agents, performing tasks ranging from simple database queries to complex multi-step workflows requiring reasoning about system interactions.

Our contributions are:

\begin{enumerate}
    \item \textbf{Empirical RL environment evaluation}: We evaluated frontier models on 150 workplace tasks, revealing substantial performance gaps even among state-of-the-art models.
    
    \item \textbf{Hierarchy of agentic capabilities}: Through systematic failure analysis, we identify five hierarchical capability levels: tool use, planning, adaptability, groundedness, and common-sense reasoning. Model failures cluster predictably along this hierarchy.
    
    \item \textbf{Task-centric RL environment design}: We describe a principled approach to environment construction that emphasizes task diversity, contributions from real domain experts, and a modular architecture that supports both training and evaluation.
    
    \item \textbf{Detailed failure taxonomy}: We provide concrete examples of failure modes at each hierarchy level, offering useful insights for researchers and model developers.
\end{enumerate}

Our findings reveal that even the best-performing models fail approximately 40\% of tasks. More importantly, the \emph{nature} of failures differs systematically: weaker models fail at basic tool use and planning, while stronger models struggle primarily with common-sense reasoning requiring contextual inference beyond explicit instructions. These results contextualize some recent findings that production agents typically execute at most ten steps before requiring human intervention \citep{pan2025map}, suggesting that current deployment constraints reflect genuine capability limitations.

\section{Related Work}

\subsection{Agent Benchmarks and Evaluation}

Agent evaluation has evolved substantially from early web interaction benchmarks to comprehensive real-world assessments. MiniWoB++ \citep{liu2018reinforcement} established foundational paradigms for the evaluation of web-based agents with simplified browser tasks. \textbf{WebArena} \citep{zhou2024webarena} introduced self-hosted functional websites that incorporate tasks across e-commerce, social forums, and collaborative software. \textbf{VisualWebArena} \citep{koh2024visualwebarena} extended this to multimodal settings with tasks requiring visual comprehension.

\textbf{OSWorld} \citep{xie2024osworld} extended evaluation to full operating system interaction across Ubuntu, Windows, and macOS, with tasks that range from office productivity, system utilities, and creative applications. The benchmark identified GUI grounding and operational knowledge as primary failure modes.

The \textbf{GAIA benchmark} \citep{mialon2024gaia} took a different approach, designing questions conceptually simple for humans but challenging for AI, requiring real-world interaction, multi-modal reasoning, and web browsing.

For software engineering, \textbf{SWE-bench} \citep{jimenez2024swe} evaluates agents on 2,294 real GitHub issues that require bug localization and patch generation. \textbf{Terminal-Bench} \citep{terminalbench2025} evaluates the capabilities of command-line agents in tasks that span scientific workflows, network configuration, cybersecurity, and data analysis, each running in isolated Docker environments.

\textbf{$\tau$-bench} \citep{yao2024taubench} uses the pass@k reliability metric that measures the success rate over $k$ independent trials rather than best-of-$k$ attempts, with the results surfacing considerable reliability concerns. The follow-up $\tau^2$-bench \citep{barres2025tau2bench} extends the evaluation to dual-control environments where both the agent and the user can invoke tools.

Recent surveys \citep{yehudai2025survey,mohammadi2025survey} have proposed taxonomies organizing agent evaluation by objectives (behavior, capabilities, reliability, safety) and process (interaction modes, benchmarks, metrics). Our work complements these taxonomies by providing an empirically-derived capability hierarchy that explains \emph{why} agents fail, not just \emph{whether} they succeed.

\subsection{Production Agent Deployment}

Although academic benchmarks provide controlled evaluation settings, understanding real-world agent deployment is equally important. \citet{pan2025map} present the first large-scale systematic study of AI agents in production, surveying 306 practitioners and conducting 20 in-depth case studies in 26 domains. Their findings reveal that production agents are typically built using simple controllable approaches: 68\% execute at most 10 steps before requiring human intervention, 70\% rely on prompting off-the-shelf models instead of weight tuning, and 74\% depend primarily on human evaluation. In particular, reliability remains the top development challenge, and practitioners deliberately constrain agent autonomy to maintain operational stability.

These production patterns provide an important context for our evaluation findings. The capability gaps we identify help explain why practitioners adopt bounded autonomy and extensive human oversight: current models have not yet achieved the reliability required for extended autonomous operation.

\section{Environment Design}

\subsection{Design Philosophy}

Effective evaluation environments for LLM agents should satisfy the following requirements: (1) a coherent world model that defines the setting, organizational relationships, and domain constraints; (2) realistic entities with authentic attributes, relationships, and behaviors; and (3) a functional tool system that allows agents to perceive and act within the environment.

We adopt a \textbf{task-centric design philosophy}: tasks provide a useful training and evaluation signal, while entities and tools exist to support tasks. This principle has important implications for environment construction. Instead of solely maximizing the number of entities or tools, we optimize for realistic and challenging tasks, which in turn requires creating diverse and interconnected entities.

Complex real-world systems are not designed top-down, but evolve organically. We reflect this principle in our construction methodology: within a framework enforcing coherent relationships and properties, domain experts with professional experience populate environments with realistic entities and tasks. This organic growth approach yields data that approximates real-world complexity.

\subsection{Environment Architecture}

Our evaluation infrastructure provides six core components:

\begin{enumerate}
    \item \textbf{Sandboxed environment}: An isolated execution context for agent operation with controlled state management.
    \item \textbf{Data layer}: Rich, interconnected entities representing realistic business operations (customers, products, orders, tickets, employees).
    \item \textbf{Tool API}: Model Context Protocol (MCP) interface for structured agent-environment interaction.
    \item \textbf{Task specification}: Prompts with defined success criteria that enable automated and human evaluation.
    \item \textbf{Task management API}: Mechanisms for task assignment, completion signaling, and reward verification.
    \item \textbf{Telemetry}: Complete trajectory logging including read/write actions and final outputs.
\end{enumerate}

This architecture supports both training and evaluation, enabling the same environment to serve multiple research purposes.

\subsection{The \corecraft{} Environment}

Our RL environment simulates \corecraft{}, Inc., an online retailer of high-performance PC components. The world model encompasses:

\begin{itemize}
    \item \textbf{Customer database}: Profiles including contact information, loyalty tiers (standard, gold, platinum), purchase history, and communication preferences.
    \item \textbf{Employee database}: Profiles including contact information, department, organization structure, and permissions.
    \item \textbf{Product catalog}: PC components with specifications, compatibility constraints, pricing, and inventory status.
    \item \textbf{Order management}: Orders with line items, shipping status, payment information, and fulfillment tracking.
    \item \textbf{Support ticketing}: Tickets with priority levels, categories, and resolution status.
\end{itemize}

The e-commerce domain was selected for several reasons: it represents economically significant work where agents could be deployed with minimal incremental effort; it spans a range of task difficulties from simple lookups to complex multi-system reasoning; and customer support specifically requires the bedrock capabilities needed for real-world agency regardless of domain specifics. Notably, recent surveys indicate that finance, technology, and corporate services represent the highest-concentration deployment domains for production agents \citep{pan2025map}, with customer support among the most common application areas.

The single world model supports multiple domain-specific task sets. While this study focuses on customer support, the same \corecraft{} infrastructure can support recruiting, marketing, and social media management tasks, enabling controlled comparisons across domains.

\subsection{Tool Interface}

Agents interact through a tool system that implements the Model Context Protocol (MCP), providing structured access to search, retrieve, create, and update operations. Listing~\ref{lst:tool_schema} presents a representative tool schema.

\begin{lstlisting}[style=jsonstyle, caption={Representative MCP tool schema for customer search.}, label={lst:tool_schema}]
{
  "name": "searchCustomers",
  "description": "Search for customers matching criteria",
  "parameters": {
    "type": "object",
    "properties": {
      "name": {"type": "string"},
      "email": {"type": "string"},
      "loyaltyTier": {
        "type": "array",
        "items": {"enum": ["standard", "gold", "platinum"]}
      },
      "limit": {"type": "integer", "default": 10}
    }
  }
}
\end{lstlisting}

\subsection{Task Design}

Tasks span a range of complexity and interaction modes. Task difficulty derives from multiple sources: multi-hop reasoning requiring information synthesis across entities, domain-specific reasoning about product compatibility or policy application, and tool selection requiring judgment about which capabilities to invoke. 

Corecraft now includes both \textbf{copilot tasks}, where a human performs the task but requests agent assistance with specific subtasks, and \textbf{fully autonomous tasks}, where the agent handles the workflow end-to-end. For the purposes of this evaluation, fully autonomous tasks were used.

% We avoid tasks that test little more than a basic capability to interact with tools.
% We include both \textbf{copilot tasks}, where a human performs the task but requests agent assistance with specific subtasks, and \textbf{fully autonomous tasks}, where the agent handles the workflow end-to-end.

Simple tasks require single-step operations:

\begin{quote}
\emph{``How many refunds were there in July 2025?''}
\end{quote}

Complex tasks require multi-step reasoning and cross-system coordination:

\begin{quote}
\emph{``A customer placed an order for a gaming build but I'm getting compatibility warnings. They ordered a ZentriCore Storm 6600X CPU with a SkyForge B550M Micro motherboard, plus 32GB of HyperVolt DDR5-5600. Can you help me figure out what's wrong and suggest the cheapest fix?''}
\end{quote}

The tasks were designed by domain experts with customer support experience, ensuring realistic complexity and authentic edge cases that reflect actual workplace challenges.

\section{Experimental Setup}

\subsection{Models Evaluated}

We evaluated frontier and legacy AI models from the major AI laboratories. Our initial evaluation (November 2025) included nine models, and we subsequently evaluated newly released models in December 2025. The complete set includes the following.

\begin{itemize}
    \item \textbf{OpenAI}: GPT-5.2 (high reasoning), GPT-5, GPT-4o
    \item \textbf{Anthropic}: Claude Opus 4.5, Claude Sonnet 4.5
    \item \textbf{Google}: Gemini 3 Pro, Gemini 2.5 Pro
    \item \textbf{Amazon}: Nova 2 Pro, Nova 1 Pro
    \item \textbf{Moonshot AI}: Kimi K2 Turbo
    \item \textbf{Alibaba}: Qwen3-Max
    \item \textbf{Mistral AI}: Mistral Medium 3.1
\end{itemize}

All models were accessed through their respective APIs with default parameters. Each received identical system prompts describing the environment, available tools, and task objectives.

\subsection{Evaluation Protocol}

For each task, the models received: (1) a system prompt that establishes the role of the agent and the environment context, (2) available MCP tools with complete schemas, and (3) the description of the task as a user message. Models could execute unlimited tool calls until producing a final response. Complete trajectories were recorded.

\subsection{Evaluation Criteria}

Successful completion of the task required: correct final answer or completion of the action, evaluated with an LLM judge using detailed human written rubrics. The failed trajectories were categorized by failure mode for capability analysis.

% This evaluation approach aligns with production practices, where 74\% of deployed agents rely primarily on human-in-the-loop evaluation rather than automated metrics \citep{pan2025map}. The complexity and domain-specificity of realistic workplace tasks often preclude fully automated correctness assessment.

\section{Results}

\subsection{Overall Performance}

Figure~\ref{fig:results} presents the primary results of our evaluation in December 2025, incorporating newly released models alongside previously evaluated ones. The following findings are immediately apparent:

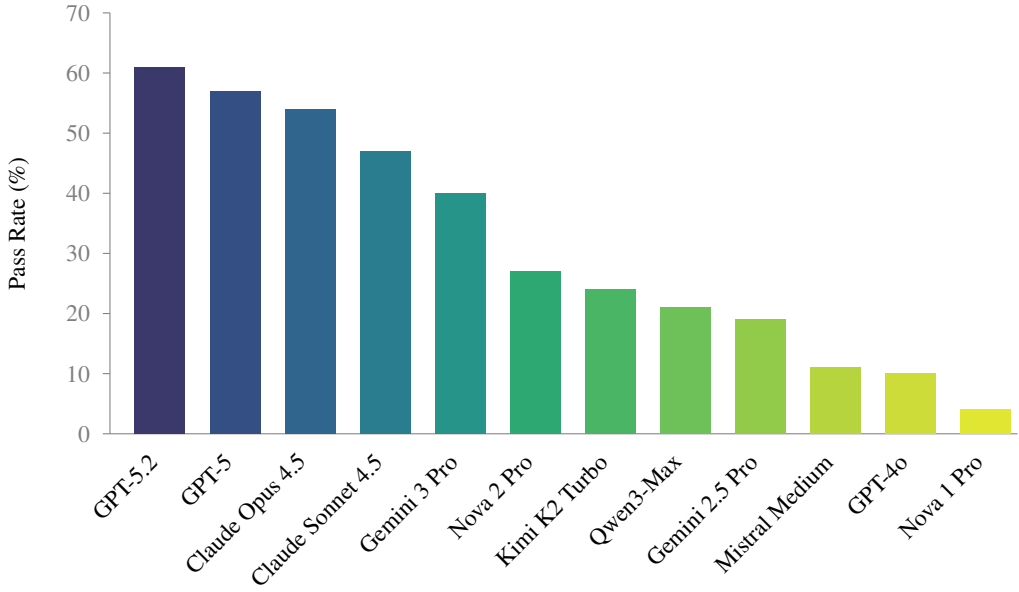
\begin{figure}[t]
\centering
% Define custom colors matching the blog post gradient (expanded to 12 colors)
\definecolor{bar1}{RGB}{59, 56, 107}    % Dark indigo
\definecolor{bar2}{RGB}{52, 79, 131}    % Deep blue
\definecolor{bar3}{RGB}{48, 102, 142}   % Steel blue
\definecolor{bar4}{RGB}{41, 125, 142}   % Teal blue
\definecolor{bar5}{RGB}{38, 148, 136}   % Dark teal
\definecolor{bar6}{RGB}{46, 168, 115}   % Sea green
\definecolor{bar7}{RGB}{74, 181, 100}   % Green
\definecolor{bar8}{RGB}{110, 193, 88}   % Light green
\definecolor{bar9}{RGB}{145, 203, 73}   % Yellow green
\definecolor{bar10}{RGB}{182, 212, 61}  % Lime
\definecolor{bar11}{RGB}{205, 220, 56}  % Yellow lime
\definecolor{bar12}{RGB}{225, 230, 50}  % Bright yellow
\begin{tikzpicture}[x=0.95cm, y=0.08cm]
    % Y-axis
    \draw[gray] (0,0) -- (0,70);
    \foreach \y in {0,10,20,30,40,50,60,70} {
        \draw[gray] (-0.1,\y) -- (0,\y);
        \node[left, font=\small, gray] at (-0.15,\y) {\y};
    }
    \node[rotate=90, anchor=south, font=\small] at (-1.0,35) {Pass Rate (\%)};
    
    % X-axis (Extended to 12.7 to fit 12 bars)
    \draw[gray] (0,0) -- (12.7,0);
    
    % Bars (width=0.7, spacing=1.05)
    \fill[bar1] (0.35,0) rectangle (1.05,61);
    \fill[bar2] (1.40,0) rectangle (2.10,57);
    \fill[bar3] (2.45,0) rectangle (3.15,54);
    \fill[bar4] (3.50,0) rectangle (4.20,47);
    \fill[bar5] (4.55,0) rectangle (5.25,40);
    \fill[bar6] (5.60,0) rectangle (6.30,27);
    \fill[bar7] (6.65,0) rectangle (7.35,24);
    \fill[bar8] (7.70,0) rectangle (8.40,21);
    \fill[bar9] (8.75,0) rectangle (9.45,19);
    \fill[bar10] (9.80,0) rectangle (10.50,11); % Mistral Medium
    \fill[bar11] (10.85,0) rectangle (11.55,10); % GPT-4o
    \fill[bar12] (11.90,0) rectangle (12.60,4);  % Nova 1 Pro
    
    % X-axis labels (rotated 45 degrees)
    \node[rotate=45, anchor=east, font=\small] at (0.70,-3) {GPT-5.2};
    \node[rotate=45, anchor=east, font=\small] at (1.75,-3) {GPT-5};
    \node[rotate=45, anchor=east, font=\small] at (2.80,-3) {Claude Opus 4.5};
    \node[rotate=45, anchor=east, font=\small] at (3.85,-3) {Claude Sonnet 4.5};
    \node[rotate=45, anchor=east, font=\small] at (4.90,-3) {Gemini 3 Pro};
    \node[rotate=45, anchor=east, font=\small] at (5.95,-3) {Nova 2 Pro};
    \node[rotate=45, anchor=east, font=\small] at (7.00,-3) {Kimi K2 Turbo};
    \node[rotate=45, anchor=east, font=\small] at (8.05,-3) {Qwen3-Max};
    \node[rotate=45, anchor=east, font=\small] at (9.10,-3) {Gemini 2.5 Pro};
    \node[rotate=45, anchor=east, font=\small] at (10.15,-3) {Mistral Medium};
    \node[rotate=45, anchor=east, font=\small] at (11.20,-3) {GPT-4o};
    \node[rotate=45, anchor=east, font=\small] at (12.25,-3) {Nova 1 Pro};
\end{tikzpicture}
\caption{Task completion rates across frontier models on 150 workplace tasks (December 2025 Update). GPT-5.2 and Claude Opus 4.5 maintain the lead, while mid-tier models like Gemini 3 Pro and Nova 2 Pro demonstrate significant progress over previous generations.}
\label{fig:results}
\end{figure}

\textbf{Finding 1}: GPT-5.2 achieves the best performance, followed by Claude Opus 4.5 and Gemini 3 Pro. These three models substantially outperform others, with a gap exceeding 10 percentage points to the next tier.

\textbf{Finding 2}: Even the best model (GPT-5.2) fails approximately 40\% of tasks. At current capability levels, autonomous agents operating without human oversight carry a fairly significant risk of error.

\subsection{The Hierarchy of Agentic Capabilities}

\begin{figure}[t]
\centering
\begin{tikzpicture}[
    level/.style={rectangle, draw, minimum width=5cm, minimum height=0.7cm, align=center, font=\small},
    arrow/.style={->, thick}
]
    \node[level, fill=red!15] (L1) at (0,0) {Level 1: Tool Use};
    \node[level, fill=orange!15] (L2) at (0,1.0) {Level 2: Planning \& Goal Formation};
    \node[level, fill=yellow!15] (L3) at (0,2.0) {Level 3: Adaptability};
    \node[level, fill=green!15] (L4) at (0,3.0) {Level 4: Groundedness};
    \node[level, fill=blue!15] (L5) at (0,4.0) {Level 5: Common-Sense Reasoning};
    
    \node[right, font=\footnotesize] at (3.0,0) {Chatbot with tool access};
    \node[right, font=\footnotesize] at (3.0,1.0) {Weak agent};
    \node[right, font=\footnotesize] at (3.0,2.0) {Weak agent};
    \node[right, font=\footnotesize] at (3.0,3.0) {Strong agent};
    \node[right, font=\footnotesize] at (3.0,4.0) {Human-level};
    
    \draw[arrow] (-3.5,0) -- (-3.5,4.4);
    \node[rotate=90, font=\footnotesize] at (-4.0,2.0) {Increasing Capability};
\end{tikzpicture}
\caption{The hierarchy of agentic capabilities. Weaker models fail at basic tool use and planning; stronger models reach common-sense reasoning as their limiting factor.}
\label{fig:hierarchy}
\end{figure}
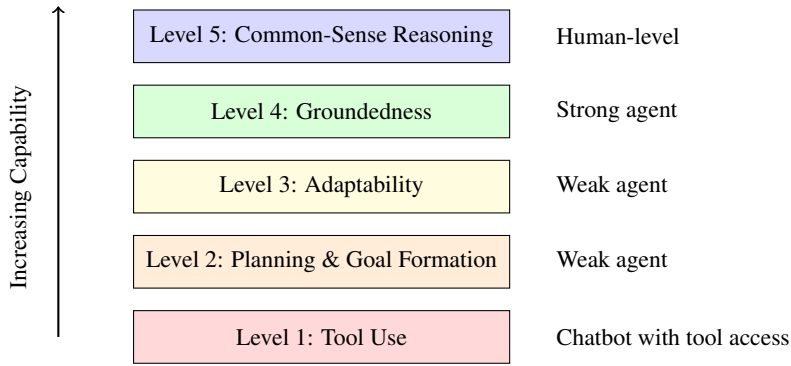

Analysis of failure trajectories revealed systematic patterns rather than random errors. Model failures clustered around specific capability levels, which we formalize as a \emph{hierarchy of agentic capabilities} (Figure~\ref{fig:hierarchy}).

The five levels are:

\begin{enumerate}
    \item \textbf{Level 1: Tool Use}. Correct invocation of tools with appropriate arguments, parsing responses, and incorporating results into reasoning.
    
    \item \textbf{Level 2: Planning and Goal Formation}. Decomposing complex tasks into subtasks, forming intermediate goals, and executing multi-step plans.
    
    \item \textbf{Level 3: Adaptability}. Recognizing when initial approaches fail and dynamically adjusting strategies based on environmental feedback.
    
    \item \textbf{Level 4: Groundedness}. Remaining anchored to the current context without hallucinating information or losing track of state across extended interactions.
    
    \item \textbf{Level 5: Common-Sense Reasoning}. Making contextually appropriate inferences beyond explicit instructions and applying world knowledge to ambiguous situations.
\end{enumerate}

This hierarchy is derived empirically, and in practice, model development is not strictly linear: these capabilities can overlap, reinforce each other, and evolve in parallel. Achieving high proficiency at a given level does not imply perfection; even frontier models can occasionally make errors in the use of basic tools. The hierarchy is best understood as a diagnostic framework for identifying where progress is solid and where foundational work remains.

\subsection{Qualitative Observations on Failure Patterns}

Weaker models failed predominantly at Levels 1--2, struggling with tool use and planning. These models rarely encountered tasks where common-sense reasoning represented the primary bottleneck, because they failed earlier in the hierarchy.

Stronger models demonstrated robust performance at lower levels, with failures concentrating at Levels 4--5. These models have largely mastered basic tool use and planning, making contextual inference the main remaining challenge.

The mid-tier models exhibited distributed failure patterns. Substantial improvements in Nova 2 Pro over its predecessor suggests that targeted training can effectively address lower-level capability gaps.

\section{Qualitative Analysis}

We present illustrative examples of failures at each capability level.

\subsection{Level 1: Tool Use Failures}

The most fundamental capability is reliable tool invocation. Weaker models frequently failed to map the prompt information to the tool arguments correctly.

\textbf{Task}: \emph{Find customers in the gold or platinum loyalty tiers who have outstanding high priority support tickets.}

Nova 1 Pro invoked \texttt{searchTickets} with the customer ID set to ``gold'' rather than using the \texttt{loyaltyTier} parameter on \texttt{searchCustomers}. This represents a fundamental failure to map task requirements to correct tool arguments.

GPT-4o correctly searched for customers in the gold and platinum tiers but made a basic mistake when searching for tickets: it passed ``high'' to the ``status'' argument rather than the explicitly available ``priority'' argument.

\subsection{Level 2: Planning Failures}

\textbf{Task}: \emph{There has been a product recall with the SkyForge X670E Pro. Please give me a bulleted list of customers who ordered this product in August 2025 with status fulfilled, paid, or pending.}

The correct workflow requires: (1) using \texttt{searchProducts} to identify the product ID, (2) using \texttt{searchOrders} to find relevant orders, and (3) returning customer names.

Both Nova 1 Pro and Mistral Medium jumped directly to \texttt{searchOrders}, passing the product name to ``product\_id''. They selected the single tool they believed would produce the final answer, then forced available data into whatever argument seemed plausible. They needed to consider all tools, determine which arguments matched the available information, and plan how to combine them.

\subsection{Level 3: Adaptability Failures}

Adaptability sharply distinguished model performance tiers.

\textbf{Task}: \emph{Hi, this is Penny Whitcomb, I am looking to upgrade my graphics card and usually go with Vortex Labs, Could you check whether the RX820L or RX780 would be compatible with parts from my last order and let me know my pricing for each?}

Gemini 2.5 Flash, Gemini 2.5 Pro, and Qwen3-Max all made the correct sequence of tool calls. However, when searching for graphics cards, they encountered a problem: they searched with \texttt{brand: "Vortex Labs"} (with space), while the database stored it as \texttt{"VortexLabs"} (without space). When searches returned empty results, they took this at face value and reported that those graphics cards were not carried.

Claude Sonnet 4.5's approach: Upon receiving empty results, Claude explicitly reasoned about the unexpected outcome and attempted alternative strategies, searching by product name patterns instead of relying solely on the brand filter. This adaptive behavior was consistently present in stronger models and absent in weaker ones.

Mid-tier models often executed a strict sequence of tool calls, but failed to adjust their plans on-the-fly when encountering unexpected results. Stronger models are capable of crafting and following multi-step plans, but also adjusting those plans mid-workflow to adapt to new information

\subsection{Level 4: Groundedness Failures}

Groundedness refers to maintaining accurate state tracking and avoiding hallucination.

\textbf{System prompt}: \emph{``The current date is September 15, 2025.''}

\textbf{Task}: \emph{Find orders from August 25--31, 2025.}

Despite the explicit system prompt, Kimi K2 Turbo searched for orders from August 25--31, \emph{2024}. When providing the final response, the year was specified as 2025. This temporal confusion represents a failure of groundedness.

Claude Sonnet 4.5 also exhibited groundedness issues. In one instance, after finding a relevant order, Claude attempted to search for customer details using a fabricated email address. When this failed, Claude self-corrected using alternative parameters, demonstrating that adaptability can partially compensate for groundedness lapses.

In another example, Claude was asked to find support tickets and report priority levels. After correctly querying for ``normal'' priority tickets, Claude's response incorrectly listed some as ``high priority'' while including them in the ``normal'' section. The response was inconsistent with the context retrieved and internally incoherent.

\subsection{Level 5: Common-Sense Reasoning Failures}

Common-sense reasoning represents the main bottleneck for frontier models.

\textbf{Task}: \emph{Identify which support tickets currently categorized as ``other'' should be reclassified as ``returns.''}

GPT-5 correctly retrieved all relevant tickets, including one stating:

\begin{quote}
\emph{``Hi, I'm really sorry but I'm going to need a refund. My son took my card without permission and made this purchase. I didn't realize until the package showed up a few hours ago. I need the money ASAP.''}
\end{quote}

GPT-5 did \emph{not} flag this for reclassification. The reasoning required: the customer requests a refund (could be return or cancelation), but ``the package showed up'', which indicates that they received the item, making it unambiguously a return. GPT-5 gathered the right information, but did not connect the dots.

\textbf{Another example}: A customer message stated:

\begin{quote}
\emph{``I've been getting frame drops when gaming so I want to upgrade my GPU. What's the highest-end GPU I can get for under \$900? Provide the price and all specifications. My name under my account should be set to Sarah Kim.''}
\end{quote}

GPT-5 interpreted this as an instruction to \emph{change} the account name rather than recognizing it as an identification. A human would immediately infer that the customer is Sarah Kim, providing their name for lookup purposes. The model's literal interpretation led to attempting an unnecessary account modification.

In a task that requires the identification of ``gamer'' customers based on purchasing patterns, GPT-5 searched through all August orders one day at a time and then individually queried the product details. A more sensible strategy would first identify categories related to gaming and then search for orders containing those products. Claude used the same inefficient approach. This represents a strategic common-sense failure rather than execution failure.

\section{Discussion}

\subsection{The Capability Hierarchy as a Diagnostic Framework}

Our hierarchy provides a diagnostic framework for understanding agent limitations. Rather than treating failure as monolithic, decomposition by capability level reveals actionable insights.

\textbf{Development priorities}: A model failing primarily in tool use requires different interventions than a model failing in common-sense reasoning. Tool use failures may benefit from improved function-calling training data; common-sense failures may require broader interventions.

\textbf{Training curriculum}: Capabilities can benefit from staged training approaches, ensuring robust lower-level competencies before addressing higher-level challenges.

\textbf{Deployment readiness}: Models that demonstrate proficiency at Levels 3--4 may be deployable with appropriate human oversight, even if Level 5 capabilities remain limited.

\subsection{The Common-Sense Ceiling}

Common-sense reasoning represents the current frontier challenge. Unlike tool use or planning, which can be improved through API training data and chain-of-thought scaffolding, common-sense reasoning requires:

\begin{enumerate}
    \item \textbf{World knowledge}: Understanding that returns include replacements, that customer messages often imply rather than state identity, and that email addresses can follow predictable patterns.
    
    \item \textbf{Contextual inference}: Recognizing when explicit instructions require implicit interpretation based on domain norms.
    
    \item \textbf{Ambiguity resolution}: Preferring interpretations coherent within task context over literal readings producing nonsensical outcomes.
\end{enumerate}

Whether common-sense reasoning can be trained explicitly or emerges from scale and diversity remains an open question with significant implications for agent development.

\subsection{Rapid Model Improvement}

Our December 2025 update reveals substantial improvements in newer model releases. GPT-5.2 achieves 61\% compared to GPT-5's 57\%, while Gemini 3 Pro (40\%) and Nova 2 Pro (27\%) show even larger gains over their predecessors (19\% and 4\% respectively). These improvements suggest that targeted training interventions can effectively address lower-level capability gaps, while the persistent failure rate on common-sense reasoning tasks indicates that this remains a more challenging frontier.

\subsection{Connection to Production Deployment Patterns}

Our findings illuminate why production agents adopt the constrained deployment patterns documented by \citet{pan2025map}. Their survey of 306 practitioners found that 68\% of production agents execute at most 10 steps before requiring human intervention, with 80\% of detailed case studies using structured workflows rather than open-ended autonomous planning. Our capability hierarchy helps explain these patterns:

\textbf{Bounded autonomy reflects capability limitations}: The approximately 40\% failure rate we observe among even the best models suggests that practitioners' preference for limited autonomous operation is well-founded. Constraining agents to structured workflows with frequent human checkpoints provides practical risk mitigation given current reliability levels.

\textbf{Human oversight compensates for higher-level failures}: Production systems' reliance on human-in-the-loop evaluation (74\% of deployed agents) addresses the common-sense reasoning gap we identify. Humans naturally handle the contextual inference and ambiguity resolution that challenge current models.

\textbf{Simple methods dominate because they work}: The finding that 70\% of the interviewed production teams use off-the-shelf models without fine-tuning aligns with our observation that frontier models already demonstrate competent tool use and planning. For many deployment scenarios, basic capabilities suffice when combined with appropriate constraints and human oversight.

\subsection{Practical Implications}

Our finding that even the best models fail approximately 40\% of tasks has a few practical implications.

\textbf{Human oversight remains important}: At current capability levels, the deployment of autonomous agents without human review carries a non-trivial risk of error. Task-specific deployment strategies with human escalation for ambiguous situations can be a more viable near-term approach. This aligns with production practice, where practitioners report that agents taking minutes to execute still substantially outperform human baselines while maintaining human verification \citep{pan2025map}.

\textbf{Importance of strong error recovery}: Claude's ability to self-correct after groundedness lapses suggests that robust adaptability can partially compensate for higher-level weaknesses, making it particularly valuable to optimize.

\textbf{Challenges in evaluation}: The difficulty of automated correctness assessment for realistic workplace tasks explains why 75\% of production agent teams evaluate without formal benchmarks \citep{pan2025map}, instead relying on A/B testing or direct human feedback.

% \subsection{Limitations}

\section{Future Work}
Several useful future directions emerge from our analysis:

\textbf{Quantitative failure analysis}: Future work will conduct a systematic quantitative analysis of failure distributions across capability levels, enabling precise measurement of how failures are distributed across the hierarchy.

\textbf{Decomposing common sense reasoning}: Investigating whether common sense can be decomposed into more concrete sub-skills forming a new hierarchy for the next frontier models.

\textbf{Task category analysis}: Examining how performance varies across task categories (simple lookup, multi-step retrieval, cross-entity reasoning, policy application, ambiguous situations) will identify specific capability gaps.

\textbf{Longitudinal evaluation}: As models evolve rapidly, longitudinal studies tracking performance will reveal whether the capability hierarchy remains stable or shifts.

% \textbf{Training interventions}: The hierarchy suggests potential training interventions targeting specific capability levels.

\textbf{Adversarial extensions}: Future versions will introduce adversarial challenges including poorly-documented tools, tools with intermittent failures requiring workarounds, and edge cases that test robustness under realistic operational conditions.

\textbf{Cross-domain evaluation}: Our RL environment supports multiple task domains (recruiting, marketing, social media management) within the same world model. Evaluating agents in these domains will reveal whether capability profiles are generalized or remain domain-specific.

\textbf{Bridging evaluation and deployment}: Following the methodology of \citet{pan2025map}, future work could examine how benchmark performance translates to production results, investigating whether our capability hierarchy predicts real-world deployment success.

\section{Conclusion}

We presented an empirical evaluation of frontier AI models in realistic workplace tasks, revealing a hierarchy of agentic capabilities for effective real-world deployment. Key findings:

\begin{itemize}
    \item Even the best models fail on a substantial portion of tasks, although newer releases show significant improvements over their predecessors.
    
    \item Failure modes cluster predictably by capability level, with weaker models struggling on fundamentals and stronger models limited by common-sense reasoning.
    
    \item Adaptability provides partial compensation for other capability gaps.
    
    \item The design of a task-centric environment through contributions from domain experts yields realistic complexity that exposes capability limitations while supporting both training and evaluation use cases.
\end{itemize}

The development of realistic RL environments is important for advancing AI agents toward economically useful work. Environments grounded in authentic workplace complexity, designed by domain experts and optimized for task diversity, provide the training signal and evaluation methodology needed to identify and address capability gaps. We have seen considerable progress in agentic AI, with models demonstrating coherent multi-step behavior and newer releases showing substantial improvements. However, substantial work remains to be done before achieving human-level performance in realistic workplace tasks. The capability hierarchy we identify provides a diagnostic framework for understanding current limitations and prioritizing the most useful areas for research and development.

\section*{Acknowledgements}

We thank the Surge AI workforce for their essential contributions to this research. Domain experts designed workplace tasks with complexity and realistic edge cases that reflect actual operational challenges. Annotators populated the \corecraft{} environment with coherent entities, relationships, and data that approximate real-world e-commerce systems. Expert evaluators assessed the model trajectories and categorized failure modes, enabling the systematic analysis that underlies our capability hierarchy.

\bibliographystyle{plainnat}

\end{document}